\documentclass[10pt,twocolumn,letterpaper]{article}
\usepackage{cvpr}

\usepackage{float}
\definecolor{cvprblue}{rgb}{0.21,0.49,0.74}
\usepackage[pagebackref,breaklinks,colorlinks,allcolors=cvprblue]{hyperref}
\usepackage[inkscapelatex=false]{svg}

\def\paperID{29} 
\def\confName{AI for Content Creation Workshop - CVPR 2025}
\def\confYear{2025}

\title{GenSync: A Generalized Talking Head Framework for Audio-driven Multi-Subject Lip-Sync using 3D Gaussian Splatting}

\author{
Anushka Agarwal \quad
Muhammad Yusuf Hassan \quad
Talha Chafekar \\
University of Massachusetts Amherst \\
{\tt\small anushkaagarw@umass.edu \quad mdhassan@umass.edu \quad tchafekar@umass.edu}
}

\begin{document}

\maketitle

\maketitle
\renewcommand{\thefootnote}{\fnsymbol{footnote}}
\footnotetext[1]{Accepted to the \textbf{CVPR} 2025 Workshop on AI for Content Creation (AI4CC)}
\renewcommand{\thefootnote}{\arabic{footnote}}

\begin{abstract}
We introduce GenSync, a novel framework for multi-identity lip-synced video synthesis using 3D Gaussian Splatting. Unlike most existing 3D methods that require training a new model for each identity , GenSync learns a unified network that synthesizes lip-synced videos for multiple speakers. By incorporating a Disentanglement Module, our approach separates identity-specific features from audio representations, enabling efficient multi-identity video synthesis. This design reduces computational overhead and achieves \textbf{6.8× faster training} compared to state-of-the-art models, while maintaining high lip-sync accuracy and visual quality.
\end{abstract}

\section{Introduction}
Lip-synced video generation is an essential task in audiovisual synthesis, with applications ranging from virtual avatars to realistic dubbing in film production \cite{wave2lip}. Generalized 2D methods such as Generative Adversarial Networks \cite{gan, peng2024synctalk}, Transformers \cite{sun2022masked}, and diffusion models \cite{xie2024x, mukhopadhyay2024diff2lip, guan2024talk} have been used to tackle this task, but struggled to capture the intricate 3D geometry and require a diverse set of training data.

To address these limitations, there has been a shift towards 3D based approaches, which uses 3D techniques like NeRFs\cite{nerf1, adnerf}, and 3D Gaussian Splatting \cite{kerbl20233d, cho2024gaussiantalker} to map the underlying 3D geometry. By incorporating person-specific 3D features, both NeRFs and 3D Gaussian Splatting enable the generation of more realistic and expressive facial animations, providing a significant improvement over 2D based methods \cite{3d_1}. However, the challenge lies in the computational and practical feasibility of these models, as current methods require a separate model trained for each speaker. This makes it both resource intensive and difficult to scale with the number of speakers.

Previous works have explored  multi-identity based approaches for talking head synthesis \cite{wiles2018x2face, siarohin2019first}, however, they struggle with problems such as handling monocular talking face videos since they require multiple input views, poor identity expression disentanglement, and inability to handle non-rigid facial motion. Existing works like \cite{minerf}  have tried to implement multi-identity synthesis, but they require explicit 3DMM expression constraints, which are computationally expensive and restrictive for real-time inference. On the other hand, gaussian based approaches like GaussianTalker \cite{cho2024gaussiantalker} work on an audio driven setting to eliminate the need for 3DMM fitting, but are limited to training only one identity per model.    

\begin{figure*}[h]
    \centering
    \vspace{-0mm}
    \includegraphics[width=0.9\textwidth]{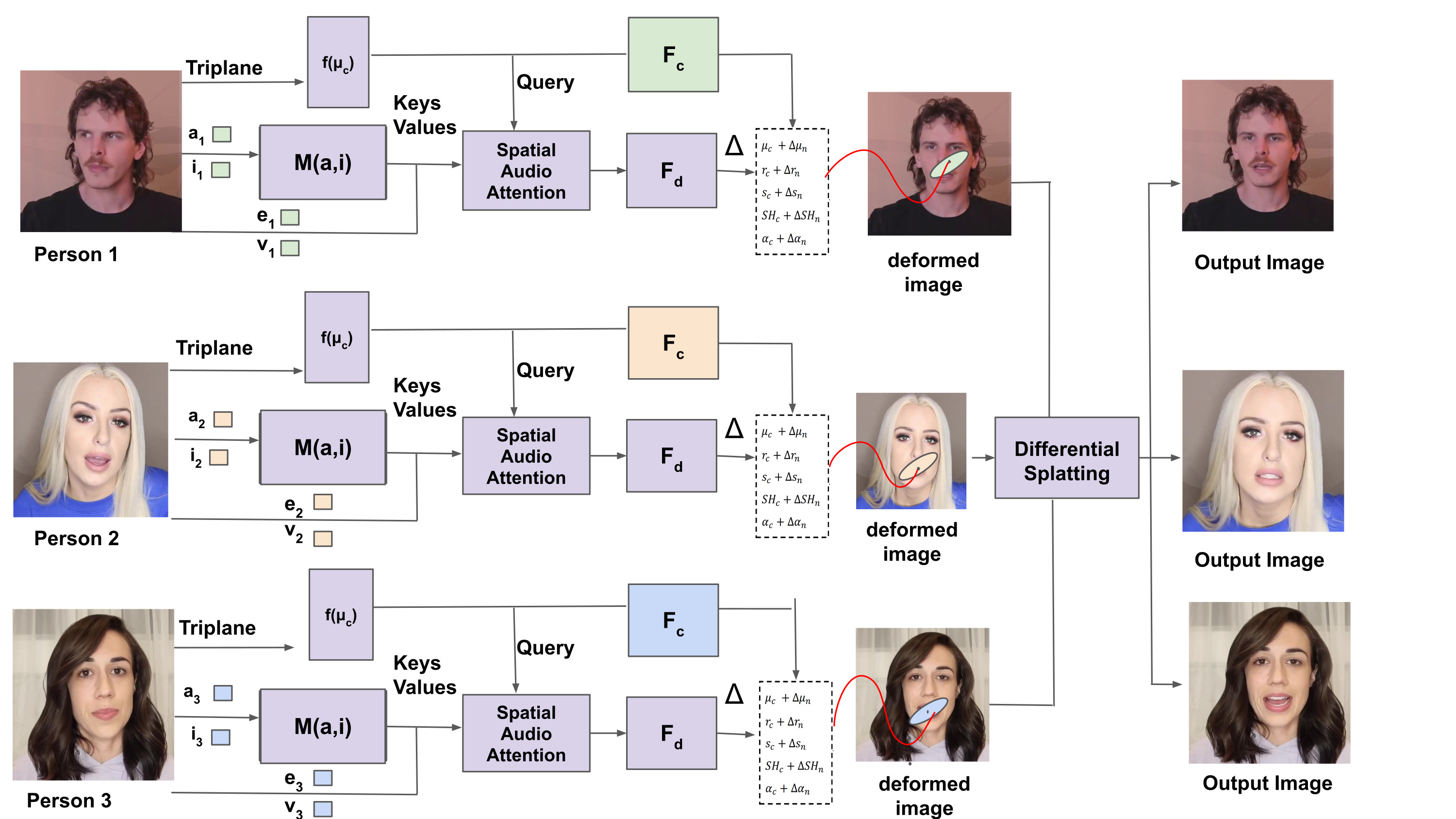}

\caption{\textbf{Overview of GenSync's pipeline.} The Deformation Module decouples representations from the audio features ($a$) and the identity vector ($i$) using a multiplicative transform (Eqn. \ref{eq:mult}). The Fused Spatial-Audio Attention Module computes cross attention among the canonical features $f(\mu_c)$ and the concatenation of the eye features ($e$), viewpoint ($v$), and the output of the Disentanglement Module. This is used to compute the deformations using an MLP $F_d$, which dynamically deforms the static canonical face output by the canonical network $F_c$ \cite{cho2024gaussiantalker}. Multiple identities are shown here for illustration purposes.}
\label{fig:final_architecture_diagram}
\vspace{-2mm}
\end{figure*}
\raggedbottom

To address this issue, we present GenSync, a generalized multi-speaker talking head module that eliminates the need for per-identity training by learning a disentangled representation between input audio and individual identity. We leverage an Identity-Aware Disentanglement Module along with Fused Spatial-Audio Attention Network that enables the model to effectively capture unique articulation styles while maintaining robust generalization capabilities across different speakers. 

\section{Methods}

Our approach follows a similar training setup to GaussianTalker~\cite{cho2024gaussiantalker} but extends it to a multi-speaker setting by learning a disentangled representation of identity and speech-driven motion across speakers. 

\subsection{Identity-Aware Disentanglement Module}
We adopt a factorized identity-audio representation inspired by MI-NeRF \cite{chatziagapi2024mi} and modify it to disentangle speaker identity from input audio. Unlike expression-driven approaches like MI-NeRF that rely on per-frame 3DMM expression extraction, which is computationally expensive and requires identity-specific optimization, we condition Gaussian-based facial motion directly on audio embeddings. This eliminates the need for per-frame optimization, enabling faster inference and real-time synthesis for capturing individual articulation styles. Our module \( M(a, i) \) (Eqn.\ref{eq:mult}), takes audio embeddings (\( a \)) and learns an identity vector (\( i \)), applying a multiplicative transformation. The Hadamard product captures non-linear dependencies between articulation dynamics and speaker identity, while additive terms preserve independent modality contributions:

\begin{equation}
\label{eq:mult}
M(a, i) = C [(U_1 a) \odot (U_2 i)] + W_2 a + W_3 i
\end{equation}

Specifically, we apply the Hadamard product between the projected audio and identity embeddings \cite{chatziagapi2024mi}, where \( U_1 \) and \( U_2 \) are learnable projection matrices that map \( a \) and \( i \) into a shared latent space. The resulting interaction is then transformed by \( C \), a learnable weight matrix. Furthermore, independent contributions from both modalities are preserved through the additive terms \( W_2 a \) and \( W_3 i \). The disentangled output of this module is then fed into the Spatial-Audio Attention module, where identity-conditioned audio features are used to refine Gaussian deformations. 

\subsection{Fused Spatial-Audio Attention Network}
The Spatial Audio Attention Module and Deformation Network\cite{cho2024gaussiantalker} together learn a fused spatial feature embedding that is both audio-aware and region specific (see Fig.\ref{fig:final_architecture_diagram}). First, the Spatial-Audio Attention Module takes the canonical features of each Gaussian $f(\mu_c)$ as query. It then computes cross-attention with the concatenation of Disentanglement Module's output along with eye features \( e \) and viewpoint \( v \). This allows the system to effectively capture how different regions of the face respond to the audio features while maintaining the influence of person-specific features.

\begin{figure*}[h]
    \centering
    \vspace{-0mm}
    \includegraphics[width=0.9\textwidth]{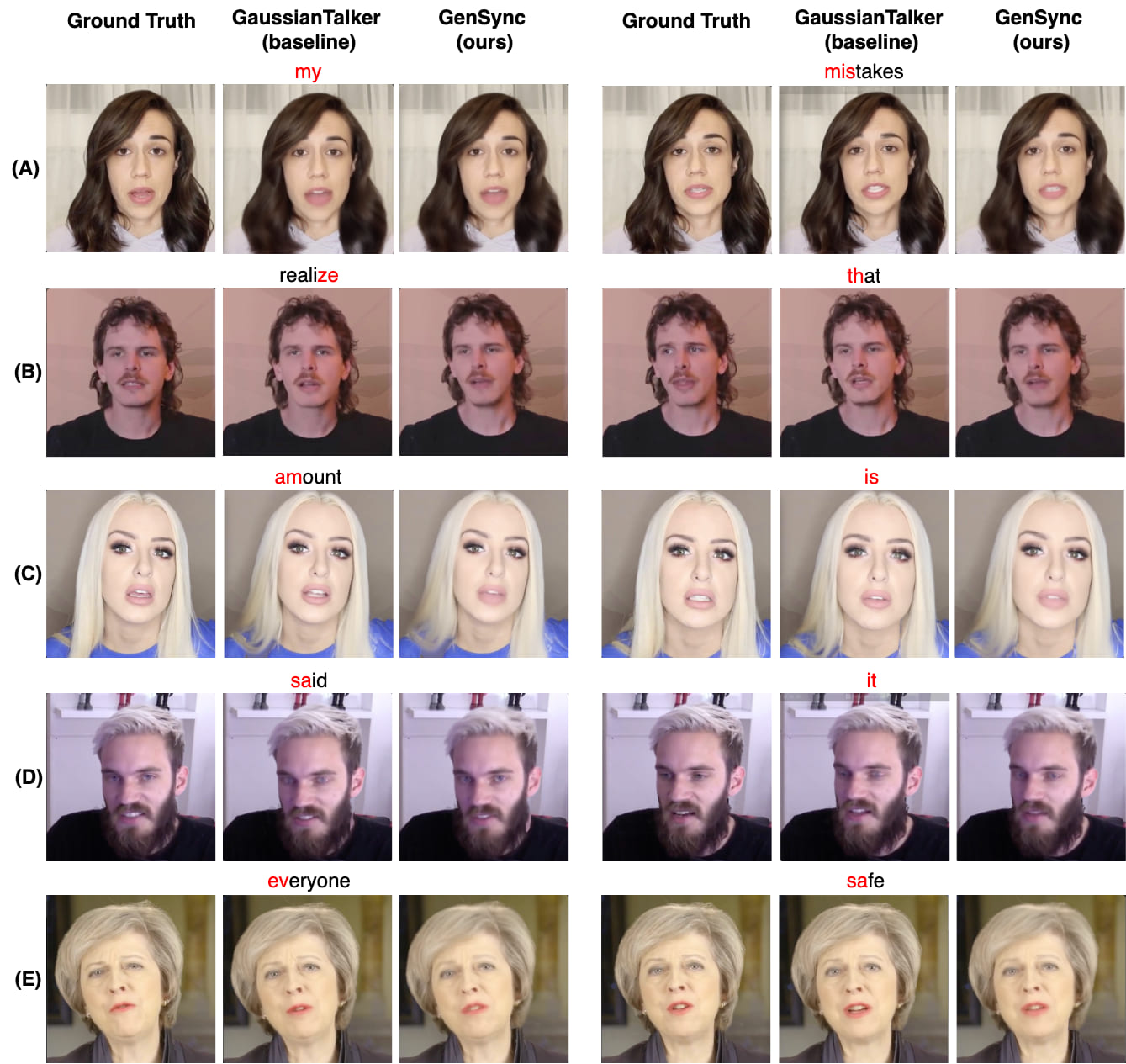}
    \caption{
    \textbf{Comparative results between GenSync (ours) and GaussianTalker (baseline) for frame-wise images from a rendered video.} The red-highlighted text indicates the current syllable being spoken. GenSync utilizes a single shared model across all identities, whereas the baseline requires separately trained models. Despite this, GenSync achieves performance comparable to GaussianTalker.
    }
    \label{fig:results_diagram}
    \vspace{-2mm}
\end{figure*}
\raggedbottom

\begin{equation}
    Q = f\mu_c, \quad
    K = V = \left[ M(a, i), e, v \right] 
\end{equation}

\begin{equation}
    h_n = \text{softmax} \left( \frac{f(\mu_c) \cdot K^\top}{\sqrt{d}} \right) \cdot V
\end{equation}

The output of the attention operation is the fused spatial-audio embedding, $h_n$, for frame $n$. This \( h_n \) is then passed through an MLP, \( F_{\text{d}} \) (Figure  \ref{fig:final_architecture_diagram}) which predicts the deformation offsets for each attribute, similar to GaussianTalker. Above, $d$ is the dimensionality of the Q, K, V vectors.

\section{Experiments and Results}

We conduct experiments on publicly available videos sourced from YouTube. We trim the video length to under 1 minute and crop each video to 512x512 dimensions. Each video is split into train frames and test frames with a ratio of 9:1. For pre-processing, we use the Basel Face Model \cite{basel} to obtain facial features and OpenFace \cite{baltruvsaitis2016openface} for eye features. We evaluate the results using LPIPS \cite{zhang2018cvpr}, FID \cite{heusel2017gans}, and SyncNet (Sync) \cite{raina2022syncnet} scores. LPIPS measures perceptual similarity using pre-trained network activations, FID computes the Fréchet distance between real and generated image distributions, and SyncNet quantifies lip-sync accuracy as the feature-wise distance between audio and video. All experiments are conducted on an NVIDIA A30 GPU. 

\subsection{Main Results}

We compare results from our model trained jointly on 10 identities with the baseline, GaussianTalker \cite{cho2024gaussiantalker}, results of which are reported in Table \ref{tab:main_results} and Figure \ref{fig:results_diagram}. For the canonical stage, we train our model for 8k iterations per identity, and then jointly train all identities in second stage for 50k iterations in total. For the baseline, we train one model for every identity and average the results across all identities.

\begin{table}[h]
\fontsize{9}{10}\selectfont
\setlength\tabcolsep{2.0pt}
\vspace{-0mm}
\centering
\vspace{-0mm}
\begin{tabular}{|l|l|l|l|l|l|l}
\toprule
                    &  LPIPS↓         & FID↓   & Sync↓ & Time↓         \\
\toprule
GaussianTalker      &  \textbf{0.073} & \textbf{20.51} & 12.26 & 62 hours \\
GenSync &  0.078          & 21.59  & \textbf{11.98} & \textbf{9 hours}        \\ \hline
\end{tabular}
\caption{Quantitative comparison of GenSync vs. GaussianTalker}

\label{tab:main_results}
\end{table}
As shown in Table \ref{tab:main_results}, our model achieves performance comparable to the baseline GaussianTalker approach in terms of LPIPS and FID scores, with minimal reductions in these metrics. In fact, our synchronization scores are slightly higher than those of the GaussianTalker baseline. Figure \ref{fig:results_diagram} further illustrates that the visual quality of generations produced by our GenSync model is on par with both the GaussianTalker baseline and the ground truth. Specifically, for Identity C, when the word amount is pronounced, the lip movements generated by our GenSync model closely resemble those in the ground truth frame. Additionally, for Identity D, we observe that the GaussianTalker approach fails to render the identity’s eyes accurately, an issue that our GenSync model successfully mitigates.

We also report the comparative training time of GenSync and GaussianTalker for 10 identities in Table \ref{tab:main_results}. GaussianTalker requires around 62 hours to train, whereas our approach achieves the same in approximately 9 hours, making it \textbf{6.8× faster}.

\begin{figure}[h]
    \centering
    \vspace{-0mm}
    \includegraphics[width=0.95\linewidth]{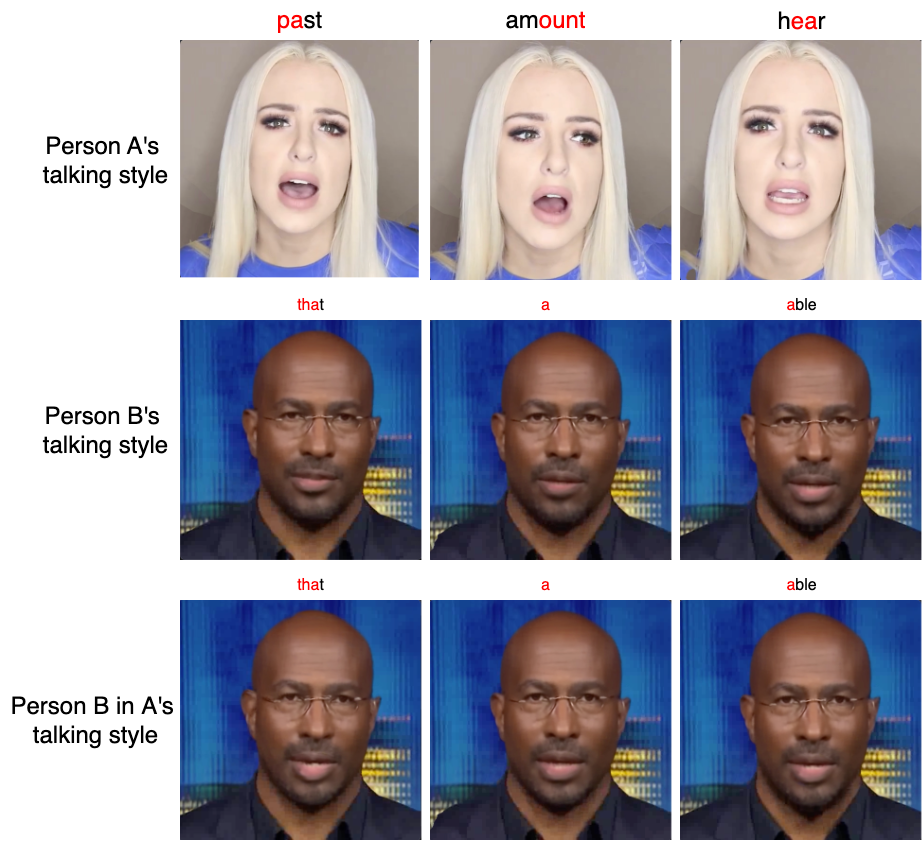}
    \caption{GenSync's output for Identity A and Identity B. We use A's learned identity embedding to generate B's video and we see the speaking style (mouth more open) transferred from A to B.}
    \label{fig:switch_ablation}
    \vspace{-2mm}
\end{figure}
\raggedbottom

\begin{figure}[h]
    \centering
    \vspace{-0mm}
    \includegraphics[width=0.95\linewidth]{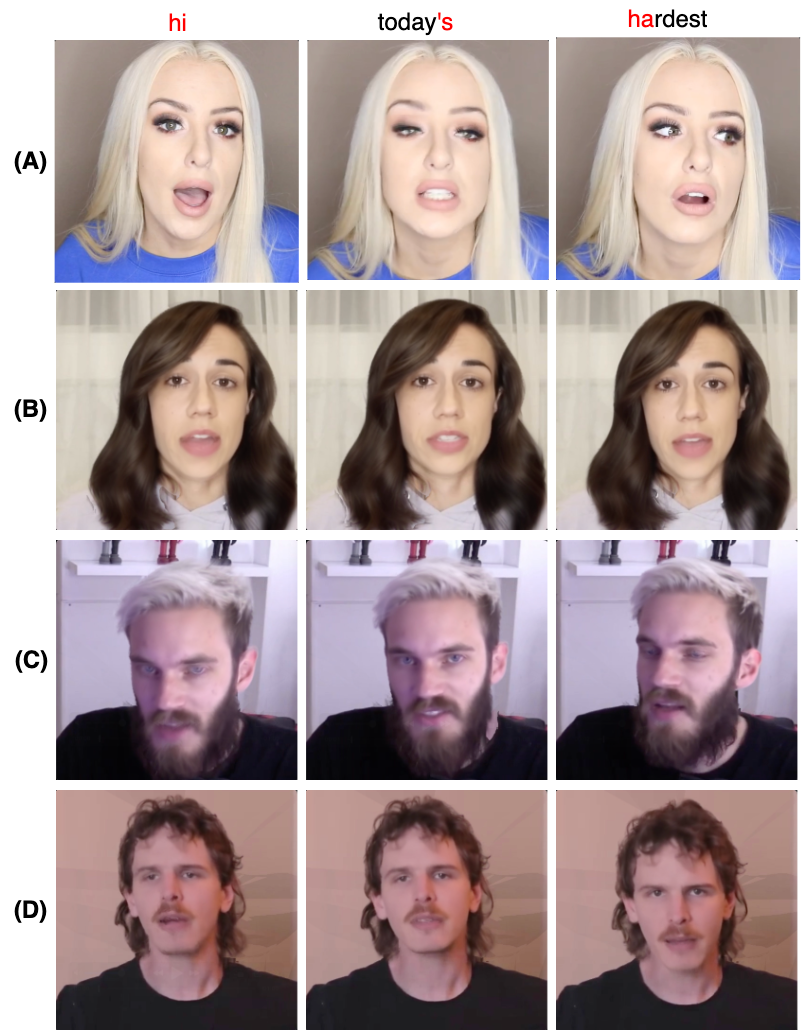}
    \caption{GenSync's output for Identity B, C and D with Identity A's audio (novel audio distribution) for the same time step. GenSync shows robustness to distribution shifts in the driving audio. For instance, the model generates plausible lip movement for male speakers C and D even when the driving audio is from female speaker A.}
    \label{fig:novel_voice}
    \vspace{-2mm}
\end{figure}
\raggedbottom

\subsection{Switching Identity Embeddings}
We conduct an experiment to analyze what the identity embedding captures. As shown in Figure \ref{fig:switch_ablation}, Identity A exhibits a speaking style characterized by a widely open mouth, whereas Identity B’s speaking style involves comparatively less mouth opening. To investigate the effect of the identity embedding, we use the learned embedding of Identity A while rendering the video of Identity B. We observe that the speaking style of A is transferred to B, resulting in B’s rendered video displaying a more open mouth compared to its original render while speaking the same words.

\subsection{Testing on Novel Audio Distributions}
We evaluate our model in a cross-driven setting where the input audio belongs to a new distribution from the one used during training. For instance, we use Identity A's audio and generate lip movements for all other identities. As shown in Figure \ref{fig:novel_voice}, the synthesized lip movements remain plausible and align well with the spoken content in the driving audio. Furthermore, our model demonstrates robustness to significant distribution shifts in voice characteristics. Notably, despite the driving audio being from a female speaker, the model effectively generates synchronized lip movements for male speakers, as demonstrated by C and D.

\section{Conclusion}
In this work, we propose a multi-person lip-sync approach using Gaussian Splatting. Our method incorporates a Disentanglement Module to separate identity-specific features from shared deformation representations, enabling a single model to synthesize lip-synced videos for multiple identities. This design allows us to achieve up to \textbf{6.8× training speed} compared to the SOTA method, GaussianTalker. Experimental results demonstrate competitive quantitative and qualitative performance, highlighting the effectiveness of our approach. Future work could include transitioning from a two-stage to a single-stage training paradigm, which could further improve efficiency and streamline training. In addition, one could work on massively scaling up the number of speakers to enhance the generalization of our method. 

{
    \small
    \bibliographystyle{ieeenat_fullname}
    \bibliography{main}
}


\end{document}